\documentclass[10pt, twocolumn, letterpaper]{article}

\usepackage[pagenumbers]{cvpr} 
\usepackage{comment}
\usepackage{lipsum}
\usepackage{multirow}
\usepackage{float}
\usepackage[ruled,vlined]{algorithm2e} 
\usepackage{graphicx}
\usepackage{etoolbox}
\usepackage{caption}
\usepackage{svg}
\usepackage{gensymb}
\usepackage{multibib}
\newcites{supp}{Supplementary}

\usepackage{xargs}                      
\usepackage[colorinlistoftodos,textsize=tiny]{todonotes}
\newcommandx{\change}[2][1=]{\todo[linecolor=red,backgroundcolor=red!25,bordercolor=red,caption={},#1]{#2}}
\newcommandx{\info}[2][1=]{\todo[linecolor=OliveGreen,backgroundcolor=OliveGreen!25,bordercolor=OliveGreen,#1]{#2}}
\newcommandx{\improvement}[2][1=]{\todo[linecolor=Plum,backgroundcolor=Plum!25,bordercolor=Plum,#1]{#2}}
\newcommandx{\thiswillnotshow}[2][1=]{\todo[disable,#1]{#2}}

\SetAlgorithmName{Alg.}{} 

\definecolor{red}{rgb}{1, 0.7, 0.7} 
\definecolor{orange}{rgb}{1, 0.85, 0.7} 
\definecolor{yellow}{rgb}{1, 1, 0.7} 
\definecolor{orange(webcolor)}{rgb}{1.0, 0.65, 0.0}
\definecolor{darkblue}{rgb}{0.0, 0.0, 0.55}

\definecolor{tabfirst}{rgb}{1, 0.7, 0.7} 
\definecolor{tabsecond}{rgb}{1, 0.85, 0.7} 
\definecolor{tabthird}{rgb}{1, 1, 0.7} 

\newcommand{\tablefont}[0]{\fontsize{8.5pt}{8.5pt}\selectfont}

\definecolor{cvprblue}{rgb}{0.21, 0.49, 0.74}
\usepackage[pagebackref, breaklinks, colorlinks, citecolor = cvprblue]{hyperref}

\def\paperID{****} 
\def\confName{****}
\def\confYear{****}

\title{CAP4D: Creating Animatable 4D Portrait Avatars\\with Morphable Multi-View Diffusion Models}

\author{
Felix Taubner\textsuperscript{1,2}
\space\space\space\space\space\space\quad
Ruihang Zhang\textsuperscript{1}
\space\space\space\space\space\space\quad
Mathieu Tuli\textsuperscript{3}
\space\space\space\space\space\space\quad
David B. Lindell\textsuperscript{1,2}
\\
\small{\textnormal{
\textsuperscript{1}University of Toronto
\space\space\space\space\space\space 
\textsuperscript{2}Vector Institute
\space\space\space\space\space\space 
\textsuperscript{3}LG Electronics
}}
}

\begin{document}

\twocolumn[{
\maketitle
\vspace{-3em}
\begin{center}
    \url{https://felixtaubner.github.io/cap4d}
\end{center}
}]

\begin{figure*}
    \centering
    \makebox[\textwidth][c]{
    \includegraphics[width=6.9in,height=8.1in]{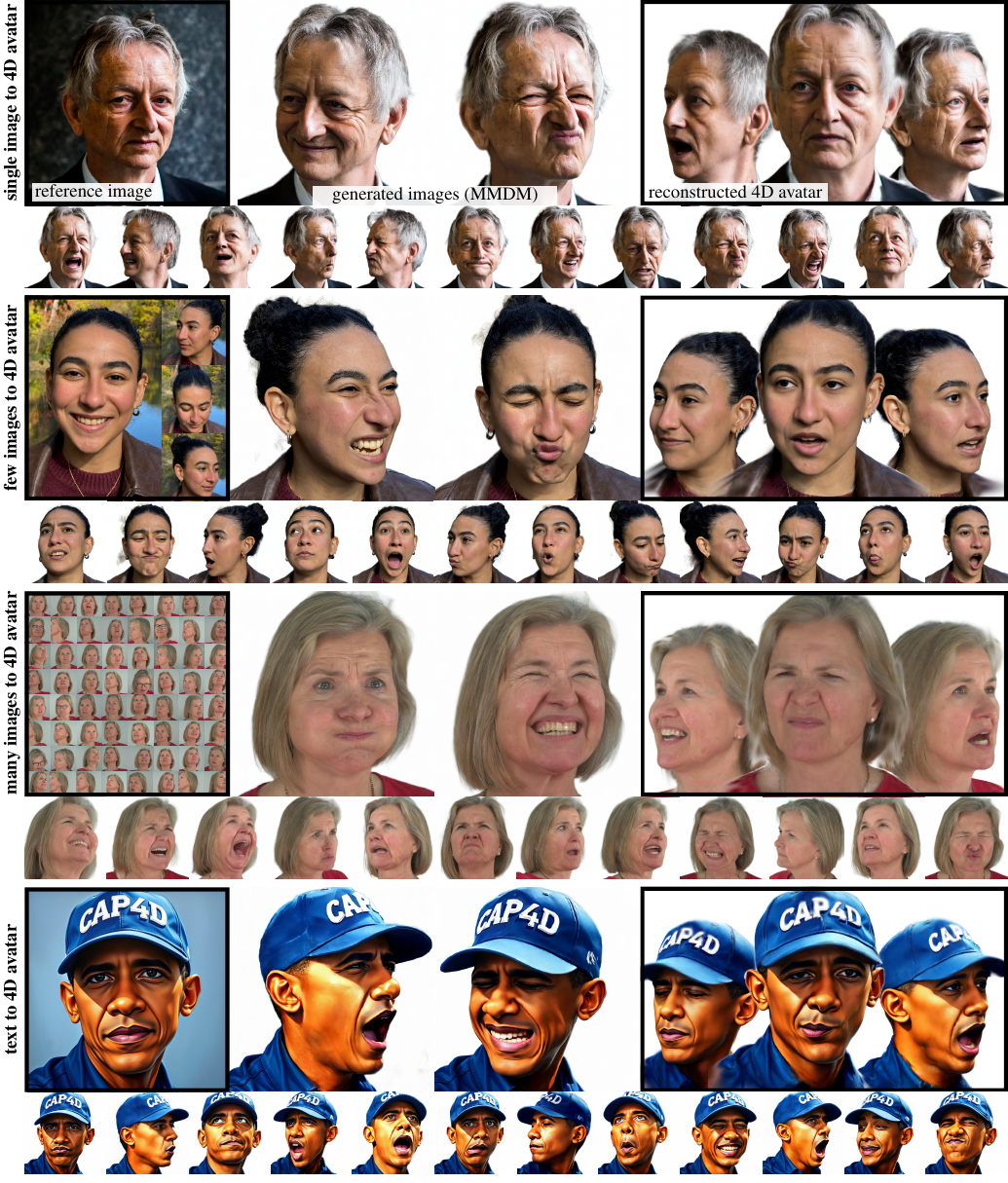}    
}
\caption{We present CAP4D: a method that generates 4D portrait avatars based on an arbitrary number of reference images (e.g., from one to one hundred) and animates them in real time. Our approach uses a morphable multi-view diffusion model to predict novel views with unseen expressions. For each subject, we generate hundreds of such views and train an animatable avatar using a representation based on 3D Gaussian splatting. Our method demonstrates state-of-the-art results for portrait view synthesis from a single image, monocular videos, or multi-view camera setups based on visual quality, identity consistency, 3D structure, and temporal consistency.}
    \label{fig:teaser}
\end{figure*}

\begin{abstract}
Reconstructing photorealistic and dynamic portrait avatars from images is essential to many applications including advertising, visual effects, and virtual reality. 
Depending on the application, avatar reconstruction involves different capture setups and constraints---for example, visual effects studios use camera arrays to capture hundreds of reference images, while content creators may seek to animate a single portrait image downloaded from the internet. 
As such, there is a large and heterogeneous ecosystem of methods for avatar reconstruction. 
Techniques based on multi-view stereo or neural rendering achieve the highest quality results, but require hundreds of reference images.
Recent generative models produce convincing avatars from a single reference image, but visual fidelity yet lags behind multi-view techniques.
Here, we present CAP4D: an approach that uses a morphable multi-view diffusion model to reconstruct photoreal 4D (dynamic 3D) portrait avatars from any number of reference images (i.e., one to 100) and animate and render them in real time. 
Our approach demonstrates state-of-the-art performance for single-, few-, and multi-image 4D portrait avatar reconstruction, and takes steps to bridge the gap in visual fidelity between single-image and multi-view reconstruction techniques.

\end{abstract}

\section{Introduction}

Reconstructing realistic human avatars from images is a sought-after capability for applications including advertising, cinema, content creation, teleconferencing, and virtual reality.
Depending on the application, avatar reconstruction involves different capture setups and constraints---from elaborate visual effects workflows involving hundreds of reference images~\cite{alexander2010digital} to more constrained settings where content creators seek to animate a single “in-the-wild” image~\cite{siarohin2019first}.
In every application, photorealism and fidelity to the subject's likeness are paramount.
In this paper, we seek a general method to reconstruct photorealistic 4D (dynamic 3D) portrait avatars that are consistent with the likeness of any number of input reference images---e.g., from one to 100---while enabling real-time animation and rendering.

Conventional methods for reconstructing photorealistic, animatable avatars rely on setups involving camera arrays~\cite{alexander2010digital,lombardi2021mixture,kirschstein2023nersemble,li2017learning} or monocular video sequences~\cite{garrido2013reconstructing,ichim2015dynamic,gafni2021dynamic,zielonka2022insta,xiang2024flashavatar}.
These setups aim to capture sufficient variation in poses and expressions to enable 4D avatar reconstruction, often through multi-view stereo~\cite{furukawa2015multi} or neural rendering techniques~\cite{tewari2022advances}.
However, these methods struggle to produce accurate results if the captured reference images lack sufficient diversity in poses or expressions.

To address this limitation, recent techniques leverage large datasets of 2D portrait images~\cite{xie2022vfhq,nagrani2017voxceleb,karras2019style} and 3D scans~\cite{yang2020facescape,dai2020statistical,zielonka2022towards,giebenhain2023learning} to train diffusion models that capture robust priors on human appearance, enabling the reconstruction of 2D~\cite{tian2024emo}, 3D~\cite{prinzler2024joker,gu2024diffportrait3d}, or 4D avatars~\cite{chen2024morphable} from a single reference image.
Still, most diffusion-based methods focus on 2D representations~\cite{ding2023diffusionrig,xie2024x,chen2024anifacediff,tian2024emo,gu2024diffportrait3d}, and inference with diffusion models is computationally expensive, which is a major obstacle to real-time rendering and animation. 
Moreover, no existing technique for 4D avatar reconstruction scales seamlessly from one to hundreds of reference images while consistently providing photorealistic results.

Here, we introduce CAP4D, a method that uses a morphable multi-view diffusion model (MMDM) to reconstruct photoreal 4D avatars that are based on any number of reference images and that are animated and rendered in real time (see Figure~\ref{fig:teaser}).
Similar to other multi-view diffusion models~\cite{shi2023mvdream,gao2024cat3d,liu2023zero,chen2024morphable}, MMDMs generate novel views of a scene based on reference images and pose conditioning. 
Our approach uses a 3D morphable model (3DMM)~\cite{blanz1999morphable,li2017learning} to provide pose and expression conditioning for the reference images~\cite{taubner2024flowface} and to control the appearance of the generated images.

CAP4D reconstructs 4D avatars in two stages.
In the first stage, the MMDM uses an iterated generation process to synthesize hundreds of images from novel viewpoints with a wide range of expressions.
While the MMDM nominally supports only a limited number of reference and generated images, we lift this restriction through a stochastic input/output (I/O) conditioning procedure inspired by recent work on view synthesis~\cite{watson2023novel} and video generation~\cite{kuang2024collaborative}. 
Specifically, at different steps of the reverse diffusion process~\cite{song2020denoising}, we condition the diffusion model on different input reference images and noisy generated outputs, enabling generation of hundreds of novel views based on a large number of reference images. 
In the second stage, we use the generated images to train a real-time 4D avatar based on 3D Gaussian splatting~\cite{qian2024gaussianavatars}. 
We augment the representation with an expression dependent appearance model to improve photorealism, and the resulting avatar can be animated and rendered in real time.
Our approach outperforms other methods for view synthesis and animation of head avatars that use one, few, or many reference images as input, and is thus relevant to a broad range of applications.

Overall we make the following contributions.
\begin{itemize}
    \item We introduce an MMDM for multi-view portrait image generation with pose and expression control, and we propose a stochastic I/O conditioning procedure to generate self-consistent portrait images given an arbitrary number of input reference images and novel viewpoints. 
    \item We develop a technique to distill generated portrait images into a 4D avatar that is animated and rendered in real time.
    \item We perform an extensive evaluation of our approach for self-reenactment and cross-identity reenactment from one or more reference images, and we demonstrate state-of-the-art results for these tasks.
\end{itemize}

\section{Related Work}
Our work is connected to methods for avatar reconstruction that use different types of input data (e.g., multi-view imagery, monocular video, or single images).

\paragraph{Monocular and multi-view avatar reconstruction.} 
Previous work reconstructs animatable 3D avatars from multi-view images or monocular video using textured mesh models~\cite{ichim2015dynamic,ma2021pixel,grassal2022neural,zheng2022avatar,cao2022authentic,athar2024bridging}, volumetric representations~\cite{zielonka2022insta,bai2023learning,giebenhain2024mononphm}, or point-based representations~\cite{zheng2023pointavatar}. 
Textured mesh models are efficient to render and can be animated using 3DMMs~\cite{blanz1999morphable}; however, they often fail to represent detailed structures like hair or teeth. 
Alternatively, volumetric representations model fine-grained appearance and produce photoreal results, but they are more computationally expensive to render~\cite{mildenhall2021nerf}.
Further, they require more sophisticated dynamics models, such as conditioning on 3DMM parameters~\cite{gafni2021dynamic, zhao2023havatar, athar2022rignerf} or learned latent codes~\cite{giebenhain2024mononphm}.
While point-based methods can be animated via deformation~\cite{zheng2023pointavatar}, they face a tradeoff between rendering efficiency and photorealism based on the number of points in the representation. 
CAP4D builds on 3D Gaussian splatting~\cite{kerbl20233d}---a hybrid between point-based and volumetric representations that represents scenes using Gaussian primitives, is efficient to optimize, and achieves photoreal reconstruction quality~\cite{xu2024gaussian,hu2024gaussianavatar,xiang2024flashavatar,shao2024splattingavatar}. 
We adapt a real-time representation based on GaussianAvatars~\cite{qian2024gaussianavatars}, which we optimize based on the output of the MMDM. 

\begin{figure*}[ht]
\centering
\includegraphics[width=\textwidth]{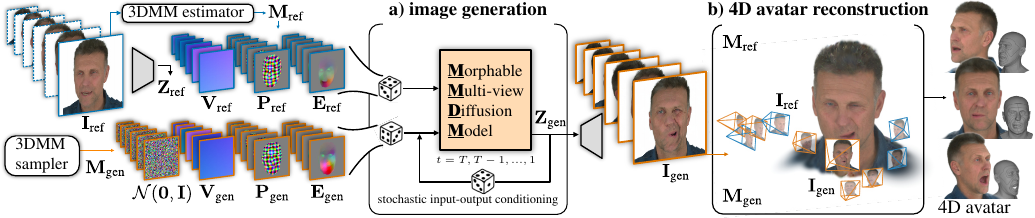}
\caption{Overview of CAP4D.  (\textbf{a}) The method takes as input an arbitrary number of reference images $\mathbf{I}_\text{ref}$ that are encoded into the latent space of a variational autoencoder~\cite{rombach2022high}. An off-the-shelf face tracker estimates a 3DMM, $\mathbf{M}_\text{ref}$, for each reference image, from which we derive conditioning signals that describe camera view direction, $\mathbf{V}_\text{ref}$, head pose $\mathbf{P}_\text{ref}$, and expression $\mathbf{E}_\text{ref}$. We associate additional conditioning signals with each input noisy latent image based on the desired generated viewpoints, poses, and expressions.  The MMDM generates images through a stochastic input--output conditioning procedure that randomly samples reference images and generated images during each step of the iterative image generation process. (\textbf{b}) The generated and reference images are used with the tracked and sampled 3DMMs to reconstruct a 4D avatar based on a 3D Gaussian splatting representation~\cite{kerbl20233d,qian2024gaussianavatars}.}
\label{fig:method}
\end{figure*}

\paragraph{Single image avatar reconstruction.}
By leveraging priors learned from large datasets, single-image methods directly regress 3D avatar representations based on textured meshes~\cite{zielonka2022towards,khakhulin2022realistic,yang2020facescape}, feature grids~\cite{ma2023otavatar,trevithick2023real,deng2024portrait4d,voodoo3d,li2024generalizable,ki2024learning}, or neural radiance fields~\cite{zhuang2022mofanerf,rebain2022lolnerf}. 
They perform novel view synthesis with expression control via 3DMMs, conditioning with latent codes, or predicted deformation fields~\cite{wang2021one,drobyshev2022megaportraits,yu2023nofa,xupv3d,yereal3d,xu2024vasa,chu2024gagavatar}.
Other approaches operate entirely in 2D, and render novel expressions and poses through learned warping and inpainting operations applied to a reference image~\cite{siarohin2019first,zakharov2019few,burkov2020neural,xu2020deep,zakharov2020fast,doukas2021headgan,ren2021pirenderer,yin2022styleheat,zhang2023metaportrait,guo2024liveportrait}.
Overall, techniques for directly regressing 3D representations require prediction in a canonical space, which can fail with extreme head poses or strong variations in appearance (e.g., non-photoreal or animated scenes). 
2D techniques often fail to inpaint discluded regions when head pose deviates significantly from the reference image.
CAP4D sidesteps the limitations of these methods by using the MMDM to generate multi-view images based on one or more reference images; then we use iterative optimization to reconstruct the 4D avatar.
Thus, our approach inherits the strengths of learned priors and iterative reconstruction methods. 

\paragraph{Multi-view diffusion models.}
Our work builds on rapid progress in the areas of 2D and 3D generation~\cite{hong2023lrm,esser2024scaling,brooks2024video,shi2023zero123++}.
We leverage latent diffusion models~\cite{rombach2022high}, which were developed for image and video generation~\cite{baldridge2024imagen,chen2024videocrafter2,guoanimatediff,blattmann2023align} and enable synthesis of 3D or 4D objects~\cite{pooledreamfusion,sun2023dreamcraft3d,bahmani2025tc4d,bahmani20244d} and 3D or 4D avatars~\cite{zhang2024rodinhd,prinzler2024joker,zhou2024headstudio}.  
We also build on multi-view diffusion models~\cite{shi2023mvdream} that generate multiple images of the same scene simultaneously for 3D reconstruction.
In this vein, we are inspired by CAT3D~\cite{gao2024cat3d}, which trains a multi-view diffusion model on large image datasets; at inference time, hundreds of novel views are generated from a single image to reconstruct the scene in 3D.
Our approach uses a similar generate-and-reconstruct paradigm, but we generate dynamic avatars rather than static scenes, and we enable controllable real-time rendering.
Close to our work, Chen et al.~\cite{chen2024morphable} train a morphable multi-view diffusion model conditioned on 3DMM information for single-image 3D avatar reconstruction.
Given a reference image, they generate a fixed number of novel views with controllable expressions and then directly infer a 3D representation.
However, their approach cannot generate consistent video frames and does not support real-time rendering.

\section{Method}

CAP4D consists of two main stages: (1) a morphable multi-view diffusion model that generates a large number of novel views from input reference images, and (2) an animatable 4D representation based on 3D Gaussian splatting representation that is reconstructed from the reference and generated images.
We provide an overview of CAP4D in Figure~\ref{fig:method}.

\subsection{Morphable Multi-View Diffusion Model}

We train an MMDM that takes a set of $R$ reference images, $\mathbf{I}_\text{ref} = \{\mathbf{i}_\text{ref}^{(r)}\}_{r=1}^R$, as input and outputs $G$ generated images, $\mathbf{I}_\text{gen} = \{\mathbf{i}_\text{gen}^{(g)}\}_{g=1}^G$. 
The model is conditioned on additional information including the head pose, expression, and camera view direction for each reference and generated image, given as $\mathbf{C}_\text{ref}=\{\mathbf{c}^{(r)}_\text{ref}\}_{r=1}^R$ and $\mathbf{C}_\text{gen}=\{\mathbf{c}_\text{gen}^{(g)}\}_{g=1}^G$.
In this way, the MMDM learns the joint probability of generated images:
\begin{equation}
P(\mathbf{I}_\text{gen} | \mathbf{I}_\text{ref}, \mathbf{C}_\text{ref}, \mathbf{C}_\text{gen}). 
    \label{eqn:mmdm}
\end{equation} 

\paragraph{Architecture.}
Our model is initialized from Stable Diffusion 2.1~\cite{rombach2022high}, and we adapt the architecture for multi-view generation following previous work~\cite{gao2024cat3d}.
Specifically, we use a pre-trained image auto-encoder~\cite{rombach2022high} to encode images into a low-resolution latent space, and we use the latent diffusion model to processes $R$ reference latent images, $\mathbf{Z}_\text{ref}$, and $G$ generated latent images $\mathbf{Z}_\text{gen}$ in parallel. 
To share information between the processed latents for each image, we replace 2D attention layers after 2D residual blocks with 3D attention (i.e., two spatial dimensions and one dimension across input images).
We also remove the cross-attention layers since we do not use a text conditioning input. 
We fine-tune the model by optimizing all parameters.

The model is conditioned on additional images that provide the head pose, expression, camera view and other contextual information for each reference and generated image.
These conditioning images consist of (1) \textit{3D pose maps}, $\mathbf{P}_\text{ref/gen}$, that provide the rasterized canonical 3D coordinates of the head geometry; (2) \textit{expression deformation maps}, $\mathbf{E}_\text{ref/gen}$, containing the rasterized 3D deformations of the geometry relative to the neutral expression mesh; (3) \textit{view direction maps}, $\mathbf{V}_\text{ref/gen}$, showing the direction of each camera ray in the first camera reference frame; 
and (4) binary masks $\mathbf{B}_\text{ref/gen}$ that indicate whether the input is a reference or generated image. 
We express the conditioning information for the reference images as $\mathbf{C}_\text{ref} = \{\mathbf{P}_\text{ref}, \mathbf{E}_\text{ref}, \mathbf{V}_\text{ref}, \mathbf{B}_\text{ref} \}$ (defined analogously for the generated images), and we concatenate them to the latent reference images, $\mathbf{Z}_\text{ref}$, as input to the network.

\paragraph{3D pose map conditioning.}
\label{sec:conditioning}
To obtain the 3D pose maps, $\mathbf{P}_\text{ref/gen}$ (illustrated in Figure~\ref{fig:method}), we use an off-the-shelf head tracker~\cite{taubner2024flowface} that jointly fits a FLAME model~\cite{li2017learning} to each reference image. 
The tracker provides the shape, head pose, and expression blendshapes, along with camera intrinsics and extrinsics.
We apply the blendshapes to a template model, $\mathbf{T}$, to recover the 3D models, $\mathbf{M}_\text{ref}=\{\mathbf{m}^{(r)}_\text{ref}\}_{r=1}^{R}$, corresponding to each reference image; 
we similarly define 3D models, $\mathbf{M}_\text{gen}=\{\mathbf{m}^{(g)}_\text{gen}\}_{g=1}^{G}$, for the generated images based on the desired head poses, expressions, and camera positions. 
Finally, we assign a texture to each vertex of $\mathbf{M}_\text{ref/gen}$, consisting of the 3D position of the corresponding vertex in the template mesh $\mathbf{T}$.

The 3D pose map is rendered by rasterizing the textures of $\mathbf{M}_\text{ref/gen}$ from the viewpoint of each reference and generated image:
\begin{equation}
\mathbf{p}_\text{ref}^{(r)} = \gamma\left[\textsc{Rasterize}\left(\mathbf{m}_\text{ref}^{(r)}, \mathbf{T}, \mathbf{\Pi}_\text{ref}^{(r)}\right)\right],
\label{eqn:3dposemap}
\end{equation}
where $\mathbf{p}_\text{ref}^{(r)} \in \mathbf{P}_\text{ref}$ is the 3D pose map for the $r$th reference image, \textsc{Rasterize} performs rasterization of the reference mesh using the associated 3D vertex position textures from the template mesh, and $\mathbf{\Pi}_\text{ref}^{(r)}$ is the camera projection matrix given by the intrinsics and extrinsics. 
The function $\gamma$ performs positional encoding~\cite{mildenhall2021nerf} that maps the rasterized 3D vertex position at each pixel into a high-dimensional feature using sine and cosine functions (see the supplement for additional details). 
We render the 3D pose maps for the generated images in the same fashion.

\paragraph{Expression deformation map conditioning.}
To facilitate the generation of subtle expression changes, we explicitly condition the network with expression deformation maps, $\mathbf{E}_\text{ref/gen}$.
We employ a procedure similar to that used for the 3D pose map, but we assign a different texture to each vertex of $\mathbf{M}_\text{ref/gen}$.
Specifically, at each vertex, we calculate the 3D offset to the corresponding vertex of a 3D model that shares the same shape blendshapes, but uses the neutral expression blendshape.
Then, we rasterize these vertex textures from the camera viewpoints of the reference and generated images. 
We omit the positional encoding step because the expression deformations have relatively low spatial frequencies~\cite{tancik2020fourier}.

\paragraph{View direction map and mask conditioning.}
For each reference and generated image, we encode the corresponding per-pixel ray directions into images, $\mathbf{V}_\text{ref/gen}$. 
We use ray directions, expressed relative to the reference frame of the first view, based on the estimated camera intrinsics and extrinsics from the tracker. 
An additional binary mask indicates whether the input image is a reference or generated image, and an outcropping mask identifies padded regions added to the reference images after center cropping around the head (see the supplement for additional details).
All conditioning images are rendered at the latent image resolution and concatenated to the reference and generated latent images before input to the MMDM.

\subsection{Generation}

The first stage of our 4D avatar reconstruction procedure is an iterative image generation process.  
Given any number of reference images as input, we generate hundreds of novel views with a range of expressions.

\paragraph{Inference with stochastic I/O conditioning.}
The appearance of occluded head regions and expression-dependent features is ambiguous if only a few reference images are provided (e.g., hair on the back of the head, teeth covered by lips, wrinkles, etc.). 
Since the MMDM architecture can only take up to four reference images as input in a single forward pass, outputs of the model when using different reference images could have a very different likeness.  
To bypass this issue, we use a stochastic input--output (I/O) conditioning procedure where we pass a random subset of input reference images and generated images to the model at each diffusion timestep. 
This procedure has multiple benefits: (1) it improves the consistency of generated images; (2) it provides a mechanism to condition on tens to hundreds of reference images; and (3) it likewise enables generating hundreds of consistent output images.

A detailed description of inference using stochastic I/O conditioning is provided in Algorithm~\ref{alg:multiplexed_ddim}.
We build on conventional denoising diffusion implicit model sampling~\cite{song2020denoising} by adding an inner loop in each diffusion timestep where we shuffle the generated images and iterate over them in batches. 
Within this inner loop, we sample a batch of the generated images and a random subset of the reference images.
Then, the model predicts the denoised generated latent images at the subsequent diffusion timestep using DDIM sampling.
After iterating through all the generated images, we proceed to the next diffusion timestep, proceeding until all images have been completely denoised.
Given a sufficient number of diffusion steps (we use 250), all reference and generated images participate jointly in the image generation process.

\begin{algorithm}[t]
\caption{Inference with Stochastic I/O Conditioning}\label{alg:multiplexed_ddim}
\KwIn{Reference image latents and conditioning
$\mathbf{Z}_\text{ref}$, $\mathbf{C}_\text{ref}$, $\mathbf{C}_\text{gen}$ \\ 
\hspace{2.6em} $R = |\mathbf{Z}_\text{ref}| = |\mathbf{C}_\text{ref}|,\; G = |\mathbf{C}_\text{gen}|$\\
\hspace{2.6em} $G'$: generated latents in each forward pass}
\KwOut{Generated image latents 
$\mathbf{Z}_{\text{gen}}$
}

\small $\mathbf{Z}_{\text{gen},T} \sim \mathcal{N}(\mathbf{0}, \mathbf{I})$ \tcp{sample noisy latents}

 \For {$t$ \text{in} ($T$,\,$T-1\,,\ldots,\,1$)}{
    \tcp{shuffle generated latents}
    \small $(\mathbf{Z}'_\text{gen,t}, \mathbf{C}'_\text{gen}) \gets \textsc{Shuffle}(\mathbf{Z}_\text{gen,t}, \mathbf{C}_\text{gen})$\\
    \For {$i$ \text{in} ($0,\,\ldots,\, G-1$)}{
        \tcp{sample w/o replacement}
        \small $(\mathbf{Z}'_\text{ref}, \mathbf{C}'_\text{ref}) \gets \textsc{RandSample}\left((\mathbf{Z}_\text{ref}, \mathbf{C}_\text{ref})\right)$\\
        \tcp{sample next batch}
        \small $(\mathbf{Z}'_\text{gen,t}, \mathbf{C}'_\text{gen}) \gets (\mathbf{Z}_\text{gen,t}, \mathbf{C}_\text{gen})[iG' +1 : (i+1)G']$\\
        \tcp{predict noise}
        \small $\epsilon_{\text{idx},t}=\textsc{mmdm}( \mathbf{Z}'_{\text{gen},t} | \mathbf{Z}'_\text{ref}, \mathbf{C}'_\text{ref}, \mathbf{C}'_\text{gen})$ \\
        \tcp{apply DDIM step~\cite{song2020denoising}}
        \small $\mathbf{Z}'_{\text{gen},t-1}=$\scriptsize$\sqrt{\alpha_{t-1}} (\frac{\mathbf{Z}_{\text{gen},t}-\sqrt{1-\alpha_t} \epsilon_{\text{idx},t}}{\sqrt{\alpha_t}}) + \sqrt{1-\alpha_{t-1}} \cdot \epsilon_{\text{idx},t}$
    }
}
\Return $\mathbf{Z}_{\text{gen}} := \mathbf{Z}_{\text{gen}, 0}$ 
\end{algorithm}

\subsection{Robust 4D Avatar Reconstruction} 

Given the reference images, generated images, FLAME parameters, and camera views, we synthesize a 4D avatar.
We build our representation based on GaussianAvatars~\cite{qian2024gaussianavatars}, which uses a collection of 3D Gaussian splats attached to the triangles of a FLAME head mesh.
Each Gaussian is linked to a specific parent triangle, with deformations modeled by expression blendshapes that drive the mesh and triangle deformations.
Additional Gaussians are added during optimization by splitting the existing Gaussians and assigning the new Gaussians to the same triangle.
Different than GaussianAvatars, we remesh the FLAME head to achieve pixel-aligned vertices in UV space at $128\times128$ resolution. 
We capture fine-grained, expression-dependent deformations using a U-Net~\cite{ronneberger2015unetconv} that predicts a UV deformation map given the offsets in UV space due to the expression blendshape.
We use our modified FLAME mesh, with an upper jaw mesh and an additional lower jaw mesh, which follows the design in GaussianAvatars.
For more information please refer to the supplement.

To optimize the representation, we use the generated images alongside the sampled expression parameters, head poses, and camera poses.
Additionally, we apply Laplacian regularization on the predicted deformation map and an $L_2$ regularization on the relative deformation and rotation of every Gaussian splat. 
We include an LPIPS~\cite{zhang2018lpips} loss to improve robustness as proposed by previous work~\cite{gao2024cat3d}, where we increase $\lambda_\text{LPIPS}$ linearly with the number of iterations.
Additional details about the optimization and regularizers are included in the supplement.

\section{Implementation}

\paragraph{Training.}
We use a collection of monocular and multi-view videos to train our model: VFHQ~\cite{xie2022vfhq}, MEAD~\cite{kaisiyuan2020mead}, Ava-256~\cite{ava256} and Nersemble~\cite{kirschstein2023nersemble}. This amounts to $24.6k$ video sequences with a total number of $41.3M$ frames of 6317 diverse subjects. 
We use an off-the-shelf multi-view head tracker~\cite{taubner2024flowface} to obtain 3DMM parameters along with a gaze estimator~\cite{abdelrahman2022l2cs} to obtain the eye rotation from the video sequences.
We train the model with the AdamW~\cite{loshchilov2019adamw} optimizer, a learning rate of $10^{-4}$, and batch size 64. We train the model for $80k$ iterations with $R=1$, then train it for an additional $20k$ iterations with randomly chosen $R<=4$ for a total of $100k$ iterations. 
During training, we randomly drop-out all conditioning signals with a probability of $0.1$, and we apply a classifier-free guidance \cite{ho2022classifierfreeguidance} during inference.
Training takes 2 weeks on 8$\times$H100 GPUs. 

\vspace{-1em}
\paragraph{Sampling and 4D reconstruction.}
We generate $G=840$ images using 250 DDIM steps with stochastic I/O conditioning, which takes around 4 hours on 4$\times$RTX6000 GPUs. 
4D avatar reconstruction requires $100k$ iterations ($\approx$4 hours on a single RTX6000 GPU).
We provide additional optimization details in the supplement.

\begin{figure*}
\begin{minipage}{\textwidth}
    \includegraphics[width=\textwidth]{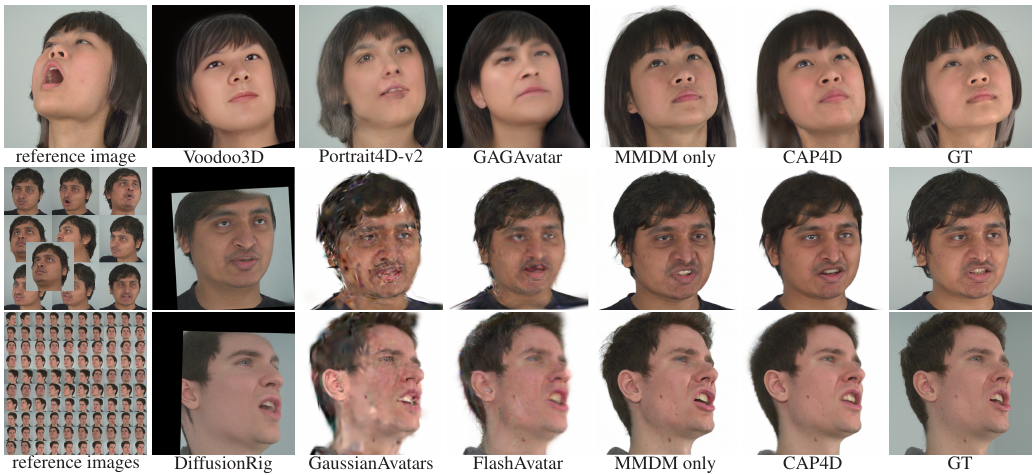}
    \vspace{-2em}
    \captionof{figure}{\textbf{Self-reenactment.} Our approach is more realistic than baseline methods for self-reenactment from a single reference image (row 1), 10 reference images (row 2) and 100 reference images (row 3). The MMDM output (MMDM only) produces the most realistic output at the cost of temporal consistency compared to our reconstructed 4D Avatar (CAP4D). }
    \label{fig:qualitative_self}
    \vspace{1em}
\end{minipage}
 
\begin{minipage}{\textwidth}
    \centering
    \begin{minipage}[t]{0.4\textwidth}
        \centering
        \tablefont
        \setlength{\tabcolsep}{3pt}
         \resizebox{0.96\textwidth}{!}{
        \begin{tabular}{ l|c c c c } 
        \toprule
        & \multicolumn{4}{c}{\textbf{single reference image}}\\\midrule
        Method & PSNR$\uparrow$ & LPIPS$\downarrow$ & CSIM$\uparrow$ & JOD$\uparrow$ \\
        \midrule
        Voodoo3D~\cite{voodoo3d} & 19.05 & 0.381 & 0.282 & 4.782 \\
        GAGAvatar~\cite{chu2024gagavatar}  & 20.78 & 0.373 & 0.457 & 5.034  \\
        Real3D~\cite{yereal3d} & 17.42 & 0.417 & 0.420 & 4.681  \\
        Portrait4D-v2~\cite{deng2024portrait4dv2}  & 16.94 & 0.404 & 0.436 & 3.871   \\
        \midrule
        MMDM only & $\textbf{21.82}$ & $\textbf{0.317}$ & 0.632 & 5.397 \\ 
        CAP4D & 21.69 & 0.311 & $\textbf{0.633}$ & $\textbf{5.672}$ \\ 
        \bottomrule
        \end{tabular}}
    \end{minipage}
    \hfill
    \begin{minipage}[t]{0.59\textwidth}
        \centering
        \tablefont
        \setlength{\tabcolsep}{1.5pt}
        \resizebox{1.0\textwidth}{!}{
        \begin{tabular}{ l |c c c c | c c c c} 
        \toprule
        & \multicolumn{4}{|c|}{\textbf{10 reference images}}  & \multicolumn{4}{c}{\textbf{100 reference images}} \\
        \midrule
        Method & PSNR$\uparrow$ & LPIPS$\downarrow$ & CSIM$\uparrow$ & JOD$\uparrow$ & PSNR$\uparrow$ & LPIPS$\downarrow$ & CSIM$\uparrow$ & JOD$\uparrow$ \\\midrule
        DiffusionRig~\cite{ding2023diffusionrig} & 16.55 & 0.450 & 0.475 & 3.89 & 16.61 & 0.446 & 0.435 &  3.86 \\
        FlashAvatar~\cite{xiang2024flashavatar} & 14.21 & 0.456 & 0.489 & 2.92 & 22.87 & 0.313 & 0.731 & 6.03 \\
        GaussianAvatars~\cite{qian2024gaussianavatars} & 18.97 & 0.448 & 0.478 & 4.33 & 20.01 & 0.416 & 0.722 & 5.10 \\
         \midrule
        no MMDM & 17.05 & 0.404 & 0.578 & 4.19 & 19.07 & 0.333 & 0.758 & 4.97 \\ 
        MMDM only & \textbf{23.82} & 0.270 & \textbf{0.804} & 6.06 & \textbf{24.12} & 0.266 & \textbf{0.803} & 6.14 \\ 
        CAP4D & 23.19 & \textbf{0.265} & 0.779 & \textbf{6.13} & 23.30 & \textbf{0.257} & 0.792 & \textbf{6.15} \\\bottomrule
\end{tabular}}
    \end{minipage}
    \vspace{-0.5em}
    \captionof{table}{\textbf{Single-image (left) and multi-image (right) self-reenactment results.} CAP4D outperforms previous methods across all metrics. Predicting images directly with the MMDM (MMDM only) trades off photometric quality (PSNR, LPIPS) and temporal consistency (JOD).}
    \label{tab:self_reenactment_combined}
    \vspace{1em}
\end{minipage}

\begin{minipage}{\textwidth}
    \includegraphics[width=\textwidth]{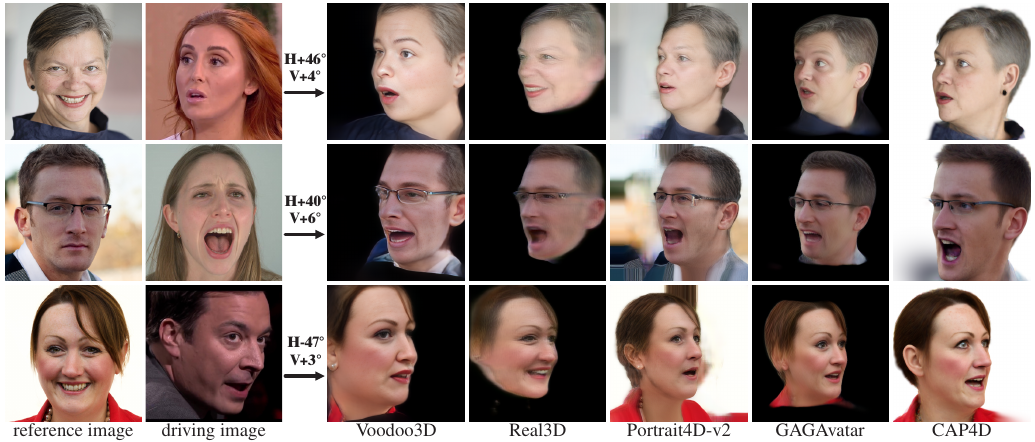}
    \vspace{-2em}
    \caption{\textbf{Cross-reenactment.} Avatars are reconstructed from a single reference image (col.\ 1), and their expressions are driven by frames of a driving video (col.\ 2). The camera moves according to the indicated horizontal (H) and vertical (V) view angle. CAP4D faithfully recovers the driving expression and maintains the likeness of the reference subject from challenging view directions. It generates plausible results in occluded regions based on the reference image (e.g., earrings, row 1) and recovers high-frequency details (freckles, row 1).  }
    \label{fig:qualitative_cross}
\end{minipage}

\end{figure*}

\begin{figure*}
    \includegraphics[width=\textwidth]{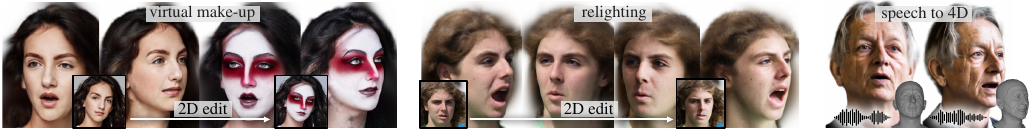}
    \caption{\textbf{Extensions.} We demonstrate 4D appearance editing and relighting by applying CAP4D to images edited using off-the-shelf models~\cite{zhang2024stablemakeup, ponglertnapakorn2023difareli}. We also animate CAP4D avatars with a method that predicts 3DMM expressions from speech~\cite{xing2023codetalker} (see supplement).}
    \label{fig:extensions}
    \vspace{-1.5em}
\end{figure*}

\section{Experiments}

We apply CAP4D to the tasks of self-reenactment and cross-reenactment and provide experimental results and comparisons to baselines. 
We also conduct an extensive set of ablation studies to assess the impact of individual components of our method: the MMDM, stochastic I/O conditioning, and the 4D representation.

\vspace{-1em}
\paragraph{Baselines}
We implement and compare our method to baselines for single-view 4D avatar reconstruction: GAGAvatar~\cite{chu2024gagavatar}, Portrait4D-v2~\cite{deng2024portrait4dv2}, Real3D-Portrait~\cite{yereal3d} Voodoo3D~\cite{voodoo3d}. 
We also include several multi-view reconstruction methods: DiffusionRig~\cite{ding2023diffusionrig},  FlashAvatar~\cite{xiang2024flashavatar} and GaussianAvatars~\cite{qian2024gaussianavatars}.
Last, we evaluate two ablated versions of our method---one without the MMDM (``w/o MMDM''; i.e., we reconstruct the avatar directly from the reference images) and one where the MMDM directly predicts the target frames (``MMDM only'').

\subsection{Self-reenactment}
We evaluate self-reenactment on nine multi-view capture sequences from the Nersemble~\cite{kirschstein2023nersemble} dataset.
We hold out 4 of the 16 camera viewpoints for evaluation (each with 100 frames).
From the remaining viewpoints, we select one (single), 10 (few), or 100 (many) reference images.
Given the reference images, we assess how well each method reenacts the appearance of the evaluation images.

CAP4D significantly outperforms every baseline in the single- and few-image reconstruction categories (\cref{tab:self_reenactment_combined}) in terms of photometric accuracy (PSNR, LPIPS), temporal consistency~\cite{mantiuk2021jod} (JOD) and identity preservation (using cosine similarity of identity embeddings~\cite{deng2022arcface}; CSIM). 
In the ``many'' reference image category, CAP4D achieves significantly higher LPIPS and CSIM than previous methods, indicating that avatars reconstructed with our method are sharper while improving identity preservation.
Although FlashAvatar achieves competitive PSNR, its output images are blurrier than CAP4D (see \cref{fig:qualitative_self}), and hence it has a lower LPIPS score. 
We find that GaussianAvatars, FlashAvatar, and our 4D avatar trained without generated images (no-MMDM) tend to overfit to the reference views and do not generalize well to novel views.
Our method improves with the number of reference images (see, e.g., CSIM), and by predicting the target frames directly (MMDM only), we achieve even higher visual quality (PSNR) at the cost of temporal consistency (JOD).
Qualitatively CAP4D produces avatars with significantly higher visual fidelity than all baselines, especially for large deviations from the reference view (e.g., \cref{fig:qualitative_self}, row 1).

\subsection{Cross-reenactment}
To evaluate cross-reenactment, we select 10 reference images from the FFHQ~\cite{karras2019ffhq} dataset.
We pair five of the images with videos from VFHQ~\cite{xie2022vfhq}, which have a normal expression range, and we select five videos with more extreme expressions from Nersemble.
The avatar is reconstructed from the reference image and the video drives its expressions.

We assess the results using the CSIM metric (\cref{tab:cross_reenactment_quanitative_main}) and in a user study, where 24 participants were presented with reference images, driving videos, and side-by-side videos generated with CAP4D and a baseline.
Then, participants indicated their preference for each of the following criteria: visual quality, expression transfer accuracy, quality of 3D structure, temporal consistency, and overall preference.

The results of the user study (\cref{tab:cross_reenactment_quanitative_main}) show a clear preference toward CAP4D overall ($74\%$ versus the most competitive baseline), as well as across all other criteria.  
Although Real3D-Portrait~\cite{yereal3d} achieves slightly better performance in the CSIM metric, human users overwhelmingly prefer CAP4D to Real3D. 
Qualitative results (\cref{fig:qualitative_cross}) indicate that CAP4D more faithfully captures the 3D appearance of the reference subject. 
Further, it preserves high-frequency detail better, is robust to large viewpoint changes, and produces 3D consistent video when other methods fail. 

\begin{table}[t]
\tablefont
\setlength{\tabcolsep}{3pt}
\begin{center}
\begin{tabular}{ l|c |c c c c | c} 
\toprule
& & \multicolumn{5}{c}{human preference} \\
Method & CSIM$\uparrow$                              & VQ & ET & 3DS & TC & Overall  \\
\midrule
Voodoo3D~\cite{voodoo3d}                   & 0.349 & 94\% & 94\% & 98\% & 99\% & 97\% \\
Real3D~\cite{yereal3d}            & \textbf{0.647} & 97\% & 90\% & 94\% & 94\% & 96\%  \\
Portrait4D-v2~\cite{deng2024portrait4dv2}  & 0.597 & 80\% & 73\% & 85\% & 89\% & 85\%  \\
GAGAvatar~\cite{chu2024gagavatar}          & 0.606 & 75\% & 63\% & 74\% & 77\% & 74\%  \\
\midrule
CAP4D                                       & 0.634 & \multicolumn{5}{c}{---} \\ 
\bottomrule
\end{tabular}
\vspace{-0.5cm}
\end{center}
\captionof{table}{\textbf{Cross-reenactment results.} We evaluate identity similarity (CSIM) and human preference based on visual quality (VQ), expression transfer (ET), 3D structure (3DS), temporal consistency (TC), and overall preference (Overall). The table reports the percentage of users (23 participants) who preferred CAP4D over the corresponding baseline in side-by-side comparisons.}
\label{tab:cross_reenactment_quanitative_main}
\vspace{-0.5cm}
\end{table}

\subsection{Ablations and Extensions}

\paragraph{Ablation study.} We investigate the design choices of our method relating to the MMDM, the stochastic sampling strategy, and the 4D reconstruction stage in \cref{tab:ablation_main}.
All ablations are conducted on the self-reenactment task with 10 reference images.
Please refer to the supplement for more ablations and qualitative comparisons.

\begin{table}[t]
\tablefont
\setlength{\tabcolsep}{3pt}
\begin{center}
\begin{tabular}{c l|c c c c} 
\toprule
Category & Ablation & PSNR $\uparrow$ & LPIPS $\downarrow$ & CSIM $\uparrow$ & JOD $\uparrow$ \\
\midrule

\multirow{3}{*}{\textbf{MMDM}} & w/o expr      & 21.64 & 0.320 & 0.669 & 5.43 \\ 
&                                w/o ray       & \textbf{22.66} & 0.315 & 0.665 & \textbf{5.59} \\
&                                Ours          & 22.54 & \textbf{0.308} & \textbf{0.668} & \textbf{5.59} \\ 

\midrule
\multirow{2}{*}{\textbf{sampling}} & w/o stochastic & 23.43 & 0.282 & 0.755 & 5.92 \\ 
&                                    Ours           & \textbf{23.82} & \textbf{0.270} & \textbf{0.779} & \textbf{6.06} \\ 

\midrule
\multirow{3}{*}{\textbf{4D rep.}} & w/o U-Net & 21.25 & 0.327 & 0.617 & 5.63 \\ 
&                                   w/o LPIPS     & \textbf{21.75} & 0.400 & 0.615 & 5.63 \\ 
&                                   Ours          & 21.69 & \textbf{0.311} & \textbf{0.633} & \textbf{5.67}  \\ 
\bottomrule
\end{tabular}
\end{center}
\vspace{-1em}
\caption{\textbf{Ablation study.} We assess the impact of removing the expression maps, view ray conditioning, stochastic I/O conditioning, and the deformation U-Net and LPIPS reconstruction loss. }
\label{tab:ablation_main}
\vspace{-0.5cm}
\end{table}

\textbf{(MMDM)} We ablate the expression deformation map (w/o expr) and view direction conditioning (w/o ray) after training the model for $30k$ steps.
We find that the expression deformation map has a significant impact on photometric quality while the impact of view direction is less significant.
\textbf{(Stochastic I/O sampling)} We directly predict the evaluation frames (MMDM-only) with and without stochastic sampling (w/o stochastic) with 10 reference frames. \cref{tab:ablation_main} shows that the stochastic sampling strategy improves the PSNR, CSIM, and JOD. 
\textbf{(4D avatar fitting)}
We ablate our U-Net, which predicts expression-dependent deformations of the 3D Gaussians (``no U-Net''); without this component we see a decrease in PSNR and LPIPS due to a reduced capability to model effects such as wrinkles. 
We also ablate the LPIPS loss---removing it improves PSNR but at a cost to LPIPS and perceptual image equality.
\vspace{-1em}
\paragraph{Extensions.} We demonstrate text-to-4D avatars generation by leveraging off-the-shelf image generation models such as Midjourney~(\cref{fig:teaser}).
Similarly, we can extend any 2D face editing model to 4D by generating avatars from an edited reference image. 
We show virtual make-up and relighting examples in \cref{fig:extensions}. 
Lastly, we exhibit speech-driven animation of a CAP4D avatar using an off-the-shelf FLAME-based~\cite{xing2023codetalker} (see videos in the supplement).

\vspace{-0.5em}
\section{Discussion}
We see multiple promising avenues for future work. 
Currently, generation is time-consuming (up to 8 hours), and while the 3DMM is convenient to animate the avatar, it does not model certain effects (e.g., tongue or hair motion). 
Future extensions could enable animation without 3DMMs and improve appearance modeling through controllable lighting (e.g., similar to Saito et al.~\cite{saito2024relightable}).
Finally, our method could be extended to model the full body. 

\vspace{-0.5em}
\paragraph{Ethics statement.} 
Digital human avatars are important to many applications, but can also be misused. We encourage responsible use of this technology (see Hancock and Bailenson~\cite{hancock2021social} for an extended discussion).

\paragraph{Acknowledgements.} 
DBL acknowledges support from LG Electronics, the Natural Sciences and Engineering Research Council of Canada (NSERC) under the RGPIN, RTI, and Alliance programs, the Canada Foundation for Innovation, and the Ontario Research Fund. The authors also acknowledge computing support provided by the Vector Institute.
{
    \small
    \bibliographystyle{ieeenat_fullname}
    \bibliography{ref}
}

\clearpage
\appendix
\setcounter{page}{1}
\maketitlesupplementary

\renewcommand{\thefigure}{S\arabic{figure}}
\renewcommand{\thetable}{S\arabic{table}} 

\setcounter{figure}{0}
\setcounter{table}{0}

This document includes supplementary implementation details and results. 
We provide implementation details related to the morphable multi-view diffusion model (MMDM), 3D morphable model (3DMM), 4D avatar, datasets, and evaluation procedures; we also provide additional evaluations and ablation studies. We encourage the reader to inspect our visual results,
comparisons with other models, and additional visualizations in the accompanying project page \href{https://felixtaubner.github.io/cap4d}{\texttt{\upshape{felixtaubner.github.io/cap4d}}}.

\section{MMDM Implementation}

\subsection{Model architecture}

\begin{figure*}
    \includegraphics[width=\textwidth]{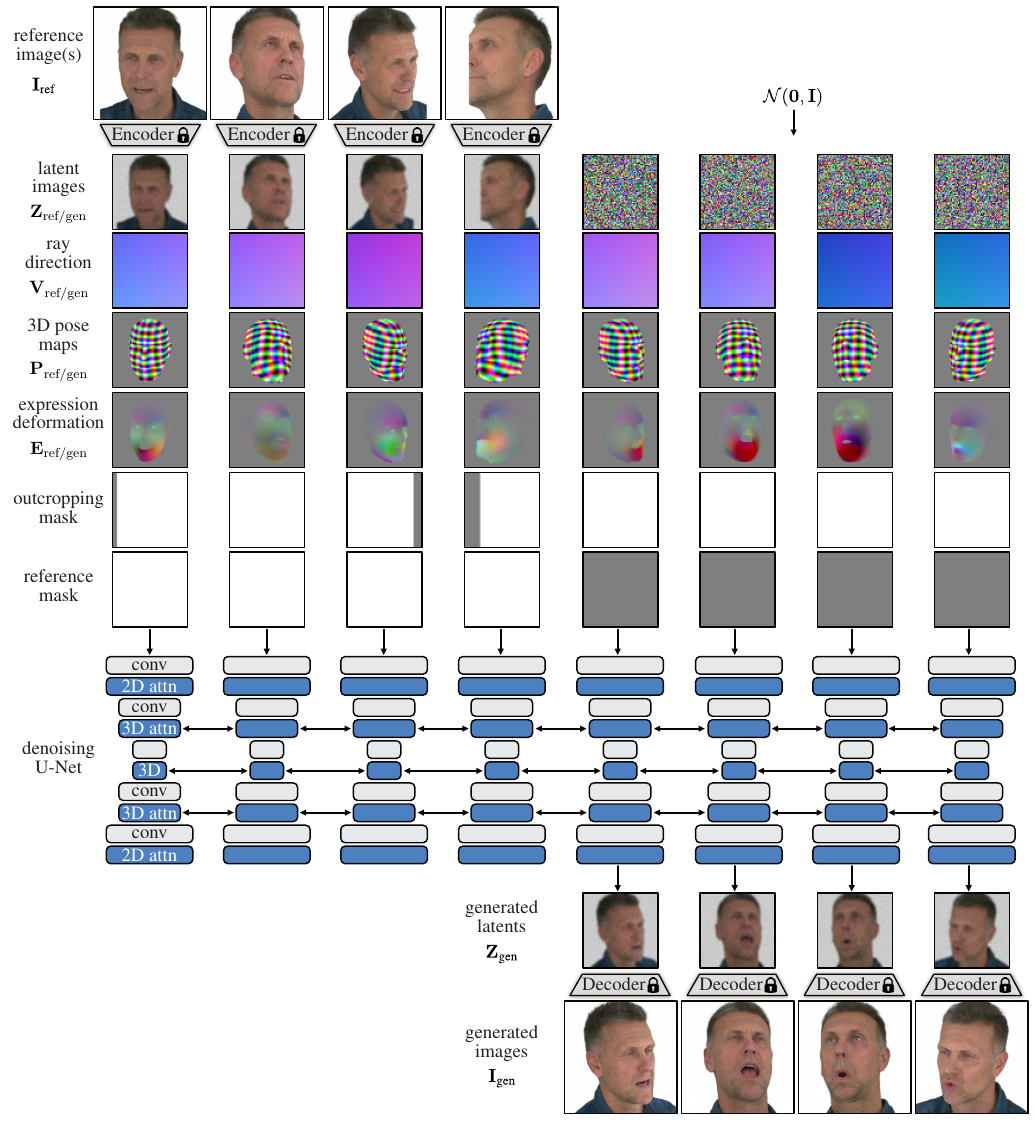}
    \caption{\textbf{MMDM architecture.} Our model is initialized from Stable Diffusion 2.1~\cite{blattmann2023stable}, and we adapt the architecture for multi-view generation following CAT3D~\cite{gao2024cat3d}. We use a pre-trained image encoder to map the input images into the latent space, and we use the latent diffusion model to process eight images in parallel. We replace 2D attention layers after 2D residual blocks with 3D attention to share information between frames. The model is conditioned using images that provide information such as head pose, expression, and camera view. The denoised latent image is decoded using a pre-trained decoder. }
    \label{fig:mmdm_architecture}
\end{figure*}

\begin{figure}
    \includegraphics[width=\linewidth]{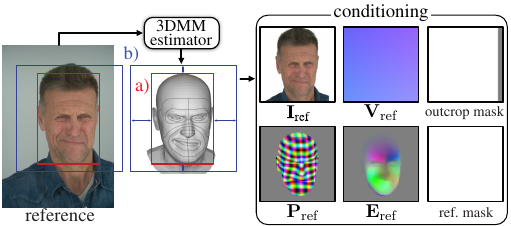}
    \caption{\textbf{MMDM conditioning.} We preprocess each reference image based on the estimated 3DMM model. We obtain a tight-fitting bounding box around the head region (a), which is squared and enlarged (b). We crop the image to the square bounding box and remove the background. Then, we update camera intrinsics to the updated crop and obtain the conditioning images $\mathbf{V}_\text{ref}$ (ray directions), $\mathbf{P}_\text{ref}$ (3D pose map), $\mathbf{E}_\text{ref}$ (expression deformation map) and outcrop mask. We follow the same process for generated images. }
    \label{fig:mmdm_conditioning}
\end{figure}

Our model is based on Stable Diffusion 2.1 \cite{blattmann2023stable} and illustrated in~\cref{fig:mmdm_architecture}. We remove all cross-attention layers and replace the 2D self-attention layers after 2D residual blocks with 3D attention layers to create the multi-view diffusion model.
Following Gao et al.~\cite{gao2024cat3d}, we only modify the 2D self-attention for layers with dimensions $32 \times 32$, $16\times16$, and $8\times8$.
We also adjust the first convolutional layer of the model to accommodate the additional conditioning channels and, where possible, initialize all layers with pre-trained weights.

During training, we update all model parameters and follow Stable Diffusion with the following adjustments.
First, we shift the signal-to-noise ratio of the noise schedule by $\log(\sqrt{N})$.
Adjusting the noise schedule provides more diffusion steps for the model to learn coarse structures in the generated images~\cite{hoogeboom2023simple, gao2024cat3d}.
Second, we found that adjusting the noise schedule to have zero terminal SNR is vital to avoid artifacts in the background~\cite{lin2024commondiffusionnoiseschedules}.
Our latent diffusion model has a total of $815M$ parameters. We use classifier free guidance weight of 2 during sampling. 

\subsection{Conditioning}
\label{sec:supp_conditioning}

The MMDM takes as input a set of reference or generated images, and each set is paired with five additional sets of conditioning images (illustrated in \cref{fig:mmdm_conditioning}): $\mathbf{V}_\text{ref/gen}$, view direction maps containing per-pixel view directions in world coordinates; $\mathbf{P}_\text{ref/gen}$, 3D pose maps 
 with rasterized vertex positions of the 3DMM template mesh; $\mathbf{E}_\text{ref/gen}$, expression deformation maps with rasterized vertex deformation vectors; and $\mathbf{B}_\text{ref/gen}$, pairs of binary masks that indicate (1) outcropped areas that are padded with white pixels and (2) whether the input is a reference or generated image.

\paragraph{Preprocessing steps and binary masks.} To create the reference conditioning images, we first obtain FLAME~\cite{li2017learning} parameters, camera intrinsic parameters, and extrinsic parameters using the 3DMM estimator of Taubner et al.~\cite{taubner2024flowface}.
To create the generated conditioning images, we sample the FLAME parameters as described in the following section.

We crop the reference images by fitting a bounding box around the vertices of the 3DMM projected onto the camera image plane.
Then, we find the smallest square bounding box that encloses the original bounding box (centered at the same location) and enlarge the result by 30\% to include the hair, neck, and shoulders; this bounding box is used for cropping.
We resize the cropped image to $512\times512$ resolution, adjust the camera intrinsics to be consistent with this cropped frame, and remove the background using an off-the-shelf background matting model~\cite{lin2021robustvideomatting}.

Sometimes, the bounding box used to crop the image extends outside the image boundaries. 
To perform outcropping in such regions, we pad the image with white pixels, and we flag these regions using a binary outcropping mask, where all outcropped areas are indicated with white pixels. 
The MMDM is conditioned on $\mathbf{B}_\text{ref/gen}$, consisting of the outcropping masks and binary masks that indicate whether the input is a reference or generated image.

\paragraph{View direction conditioning.} 
We use the camera intrinsic and extrinsic parameters to compute view conditioning images, $\mathbf{V}_\text{ref/gen}$, containing the view direction for each pixel in world coordinates.
The world coordinates are computed relative to the first reference view, which is positioned at the origin of the coordinate system with its rotation matrix set to the identity.

\paragraph{3D pose and expression deformation conditioning.} 
As described in the main text, we obtain the 3D pose map $\mathbf{P}_\text{ref/gen}$ by texturing the vertices of the tracked 3DMM model with the 3D vertex positions of the 3DMM template mesh $\mathbf{T}$. 
We rasterize these vertex positions and encode the values using a periodic positional encoding~\cite{tancik2020fourier}:
\begin{equation}
    \gamma(p)=(\sin(2^0 p),\cos(2^0 p),\dots,\sin(2^{L-1} p),\cos(2^{L-1} p)),
\end{equation}
where $p$ is the 3D vertex position texture, and $L=7$ is the number of encoding frequencies. 
This results in 42 positional encoding channels. 
We compute $\mathbf{E}_\text{ref/gen}$ in a similar fashion by rasterizing the  3D deformations caused by the expression blendshape parameters ($\mathcal{E}(\boldsymbol{\phi})$), but we omit the positional encoding. 

\subsection{MMDM sampling}

\begin{figure*}
    \includegraphics[width=\textwidth]{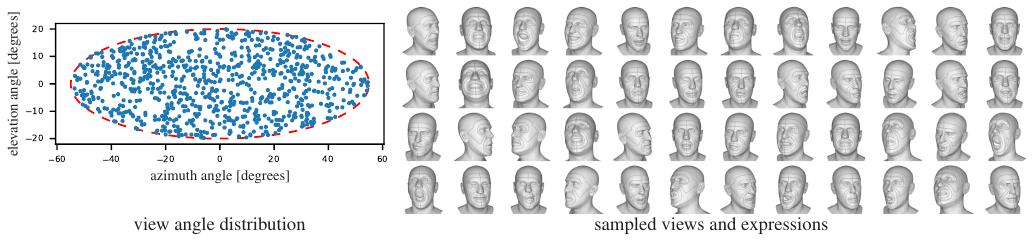}
    \caption{\textbf{3DMM Sampling.} To generate novel views we uniformly sample in azimuth and elevation (left). For each camera view, we select unique expression parameters from our expression database, which is obtained from the Nersemble dataset~\cite{kirschstein2023nersemble} following a diversity-promoting sampling scheme~\cite{salomon2013diversesampling}. A subset of the sampled expressions and views are visualized on the right.}
    \label{fig:sampling}
\end{figure*}

We follow a fixed sampling procedure to obtain novel generated views and 3DMM parameters as illustrated in~\cref{fig:sampling}.
We begin by sampling a set of $G$ generated camera views, where each view is rotated around the center of the head with a randomly sampled azimuth $\psi$ and elevation angle $\theta$ (we set the view aligned straight on with the face to have zero azimuth and zero elevation).
The camera is kept at the same distance from the head as the first reference view.  
The values $\psi$ and $\theta$ are uniformly sampled to be within an ellipse (red line, \cref{fig:sampling}):
\begin{equation}
\label{eq:periodic}
    (\frac{\psi}{\psi_\text{max}}) ^ 2 + (\frac{\theta}{\theta_\text{max}}) ^ 2 < 1,
\end{equation}
where $\psi_\text{max}=55^\circ$ and $\theta_\text{max}=20^\circ$. 

\paragraph{Expression database.}
We select a unique expression parameter for each camera view from our expression database.
The database is created using a diversity-promoting sampling scheme (implemented in the \texttt{diversipy} software package~\cite{salomon2013diversesampling}) that partitions the space of expressions obtained from all frames of the Nersemble~\cite{kirschstein2023nersemble} dataset into $G=840$ dissimilar subsets with a representative sample for each subset. 
To determine the distance between each expression sample, we use Euclidean distance in expression parameter space $\boldsymbol{\phi} \in \mathbb{R}^{65}$, where each dimension is weighted by the maximum vertex displacement of the corresponding blendshape. 

\section{FLAME 3DMM Implementation}
\label{sec:flame}

The FLAME representation \cite{li2017learning} consists of $N_\textit{v}=5023$ vertices, which are controlled by identity shape parameters $\boldsymbol{\beta}$, expression shape parameters $\boldsymbol{\phi}$, and skeletal joint poses through linear blend skinning.
We ignore the jaw pose and use the FLAME2023 model, which includes deformations due to jaw rotation within the expression blend-shapes.
Overall, there are $\boldsymbol{\beta} \in \mathbb{R}^{150}$ identity shape parameters and $\boldsymbol{\phi} \in \mathbb{R}^{65}$ expression parameters.
To model eye rotation, we use one joint rotation for both eyes.
Each vertex position is determined by adding expression and identity shape offsets to the template mesh $\mathbf{T}$, and the offsets are computed using the expression and identity shape parameters and the corresponding linear bases, $\mathcal{E}$ and $\mathcal{S}$:
\begin{equation}
    \mathbf{m} = \mathbf{T} + \mathcal{E}(\boldsymbol{\phi}) + \mathcal{S}(\boldsymbol{\beta}).
\end{equation}

We use an edited version of the FLAME template mesh to create the conditioning signals used by the MMDM. 
More precisely, we manually position a spherical mesh inside the mouth region and behind the lip to represent the upper jaw. 
This sphere is static and unaffected by the expression shape parameters $\boldsymbol{\phi}$. 
We remove the lower neck vertices and limit the conditioning model to the head region. 

For the representation used by the 4D avatar, we add a spherical mesh to model the lower jaw. 
This sphere is placed similarly to the upper jaw mesh, but it is rigged to move with the jaw joint. 
We compute the jaw rotation heuristically by tracking the deformation of a specific vertex on the lower jaw relative to the jaw joint position obtained from the FLAME model.
We adopt the UV mapping provided by previous work ~\cite{deca} and modify it manually to add textures for the upper and lower jaw meshes. 

\section{4D Avatar Implementation}

\begin{figure*}
    \includegraphics[width=\linewidth]{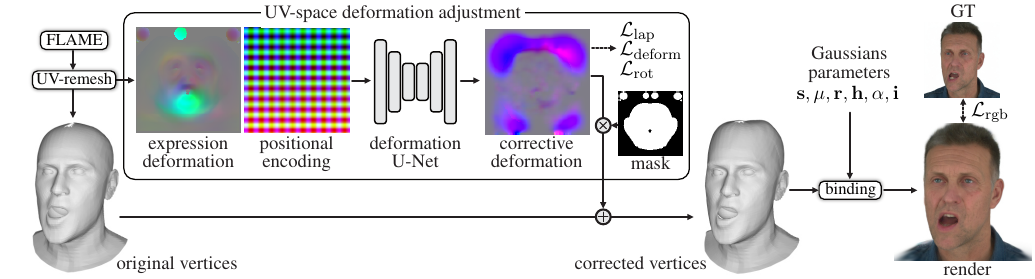}
    \caption{\textbf{Overview of 4D Avatar Model.} Our 4D representation incorporates multiple improvements to the  GaussianAvatars~\cite{qian2024gaussianavatars} model. First, we re-mesh the FLAME topology so that each vertex corresponds to a pixel in the UV space. Then, we input the UV-space deformations caused by the expression blendshapes and a UV-space positional encoding into a deformation U-Net. This U-Net outputs corrective deformations, which, after masking, are added to the remeshed FLAME output. Following Qian et al.~\cite{qian2024gaussianavatars}, the Gaussians are parameterized by a scale $\mathbf{s}$, local position $\mu$, local rotation $\mathbf{r}$, spherical harmonics coefficients $\mathbf{h}$, opacity $\alpha$, and parent triangle $\mathbf{i}$. We apply regularizers to the output of the U-Net, and we add an LPIPS penalty to the photometric loss $\mathcal{L}_\text{rgb}$. }
    \label{fig:method_4d_avatar}
\end{figure*}

Our 4D avatar model is based on GaussianAvatars~\cite{qian2024gaussianavatars} with a few modifications to make it more robust to generated views. We describe these changes in the following and illustrate the approach in \cref{fig:method_4d_avatar}.

\paragraph{Deformation model.}
We disable the per-frame fine-tuning of FLAME parameters used by GaussianAvatars during training as we find that it leads to overfitting. 
To correct inaccuracies in the underlying 3DMM, we instead use a U-Net to deform the mesh with expression-dependent deformations. 

The input to the U-Net consists of UV maps that encode the expression deformations and positional encodings of UV map pixel locations. 
To compute the expression deformation map, we first remesh the FLAME head to achieve pixel-aligned vertices in UV space at a $128 \times 128$ resolution. 
Then, we rasterize the deformations caused by the expression parameters $\mathcal{E}(\phi)$ into UV space. 
We obtain a positional encoding of the UV space by encoding the UV coordinates with the same periodic functions as in \cref{eq:periodic},
where we set the number of frequencies to $L=6$ and the coordinate $p$ to the UV-space coordinate of each pixel, leading to a total number of 24 encoding channels. 
The positional encoding is concatenated to the UV-space expression deformation and processed by a 6-layer U-Net~\cite{ronneberger2015unetconv}. 

The U-Net outputs a 3-channel deformation map, $\mathbf{D}_\text{uv}$, (see Figure~\ref{fig:method_4d_avatar}) indicating the expression-dependent deformation correction.
We mask this deformation to prevent deformations in static areas such as the back of the head and lower neck.
To obtain the final vertex positions, we add these deformations to the vertices produced by the FLAME model.

During training, we use multiple regularizers to prevent motion artifacts. 
First, we apply a weight decay of $2\times 10 ^{-3}$ on the U-Net weights; second, we use an L2 loss $\mathcal{L}_\text{lap}=||\Delta \mathbf{D}_\text{uv}||_2^2$ on the Laplacian of the deformation map; last we append an L2 loss on the relative deformation and rotation of each Gaussian $\mathcal{L}_\text{deform}$ and $\mathcal{L}_\text{rot}$.
We logarithmically decrease the learning rate of this network from $10^{-5}$ to $10^{-7}$ during training. 

\paragraph{LPIPS loss.}
To make the reconstruction more robust to inconsistencies in the generated views, we add an LPIPS~\cite{zhang2018lpips} loss to the existing photometric loss from GaussianAvatars~\cite{qian2024gaussianavatars} and weight it against the other term:
\begin{equation}
    \mathcal{L}_\text{rgb} = \lambda_\text{LPIPS} \mathcal{L}_\text{LPIPS} + (1 - \lambda_\text{LPIPS}) \mathcal{L}_\text{rgb,GA},
\end{equation}
where $\lambda_\text{LPIPS}$ is the weighting of LPIPS loss, which we linearly increase from 0 to $0.9$ during training. $\mathcal{L}_\text{rgb,GA}$ is the original photometric loss from GaussianAvatars.
We also include their scaling and positional losses $\mathcal{L}_\text{scaling}$ and $\mathcal{L}_\text{position}$, resulting in the modified total loss function:
\begin{equation}
    \mathcal{L} = \mathcal{L}_\text{rgb} + \lambda_\text{deform} \mathcal{L}_\text{deform} + \lambda_\text{rot} \mathcal{L}_\text{rot} + \mathcal{L}_\text{scaling} + \mathcal{L}_\text{position},
\end{equation}
where $\lambda_\text{deform}=0.4$ and $\lambda_\text{rot}=0.005$ are the weights for the corresponding losses. For more information on these loss functions, we refer to GaussianAvatars~\cite{qian2024gaussianavatars}.

\paragraph{Other changes.}
We attach the Gaussians to the triangles of the re-meshed FLAME model. Each Gaussian contains a scale $\mathbf{s}$, local position $\mathbf{\mu}$, local rotation $\mathbf{r}$, spherical harmonics coefficients $\mathbf{h}$, opacity $\alpha$ and parent triangle $\mathbf{i}$.
We initialize the avatar with $100K$ Gaussians, where the number of Gaussians for each triangle is proportional to the area of the triangle.
Also, we set each Gaussian's initial scale to be inversely proportional to the number of Gaussians per triangle, which we find to reduce rendering artifacts. 
\begin{table*}[t!]
\begin{minipage}{\textwidth}
    \centering
    \begin{minipage}[t]{1.0\textwidth}
        \centering
        \tablefont
        \setlength{\tabcolsep}{3pt}

        \begin{tabular}{ l|c c c c } 
                \toprule
                & \multicolumn{4}{c}{\textbf{single reference image}}\\\midrule
                Method & SSIM $\uparrow$ & AED $\downarrow$ & APD $\downarrow$ & AKD $\downarrow$ \\
                \midrule
                Voodoo3D~\cite{voodoo3d} & 0.658 & 1.463 & 0.085 & 10.52 \\
                GAGAvatar~\cite{chu2024gagavatar} & 0.718 & 1.076 & 0.069 & 13.01  \\
                Real3D~\cite{yereal3d} & 0.667 & 1.561 & 0.118 & 15.91  \\
                Portrait4D-v2~\cite{deng2024portrait4dv2}  & 0.651 & 1.291 & 0.121 & 19.97 \\
                \midrule
                MMDM only & 0.730 & \textbf{0.707} & \textbf{0.041} & 5.82 \\ 
                CAP4D & \textbf{0.748} & 0.782 & \textbf{0.041} & \textbf{5.68} \\ 
                \bottomrule
        \end{tabular}
        \vspace{2em}
    \end{minipage}
    \begin{minipage}[t]{1.0\textwidth}
        \centering
        \tablefont
        \setlength{\tabcolsep}{1.5pt}

        \begin{tabular}{ l |c c c c} 
        \toprule
        & \multicolumn{4}{|c}{\textbf{10 reference images}}\\
        \midrule
        Method & SSIM $\uparrow$ & AED $\downarrow$ & APD $\downarrow$ & AKD $\downarrow$ \\\midrule
        DiffusionRig~\cite{ding2023diffusionrig}       & 0.624 & 0.930 & 0.084 & 17.8 \\ 
        FlashAvatar~\cite{xiang2024flashavatar}        & 0.580 & 1.243 & 0.124 & 20.8 \\ 
        GaussianAvatars~\cite{qian2024gaussianavatars} & 0.628 & 1.413 & 0.237 & 21.5 \\
         \midrule
        no MMDM     & 0.590 & 1.064 & 0.093 & 10.6  \\ 
        MMDM only   & \textbf{0.753} & \textbf{0.542} & \textbf{0.033} & \textbf{5.09}  \\ 
        CAP4D       & 0.748 & 0.782 & 0.041 & 5.68  \\\bottomrule
        \end{tabular}
    \vspace{2em}
    \end{minipage}
    
    \begin{minipage}[t]{1.0\textwidth}
    \centering
    \tablefont
    \setlength{\tabcolsep}{1.5pt}
    \begin{tabular}{ l |c c c c} 
        \toprule
        & \multicolumn{4}{|c}{\textbf{100 reference images}} \\
        \midrule
        Method & SSIM $\uparrow$ & AED $\downarrow$ & APD $\downarrow$ & AKD $\downarrow$   \\\midrule
        DiffusionRig~\cite{ding2023diffusionrig} & 0.617 & 0.975 & 0.085 & 18.6 \\ 
        FlashAvatar~\cite{xiang2024flashavatar}  & 0.741 & 0.689 & 0.042 & 6.16 \\ 
        GaussianAvatars~\cite{qian2024gaussianavatars} & 0.713 & 0.737 & 0.062 & 9.01 \\
         \midrule
        no MMDM & 0.675 & 0.653 & 0.058 & 7.88 \\ 
        MMDM only & 0.754 & \textbf{0.535} & \textbf{0.032} & \textbf{5.07} \\ 
        CAP4D & \textbf{0.763} & 0.634 & 0.035 & 5.21 \\\bottomrule
    \end{tabular}
    \end{minipage}
    \captionof{table}{\textbf{Additional single-image (top) and multi-image (middle, bottom) self-reenactment metrics.} Our method consistently outperforms baselines in terms of expression accuracy (AED), photometric quality (SSIM) and alignment accuracy (AKD). }
    \label{tab:self_reenactment_combined_extra}
\end{minipage}
\end{table*}
\begin{table*}[h!]
\tablefont
\setlength{\tabcolsep}{3pt}
\begin{center}
\begin{tabular}{ l|c c } 
\toprule
Method & AED $\downarrow$ & APD $\downarrow$ \\
\midrule
Voodoo3D~\cite{voodoo3d} & 2.428 & 0.082  \\
GAGAvatar~\cite{chu2024gagavatar}  & 2.137 & 0.079  \\
Real3D~\cite{yereal3d} & 2.581 & 0.104  \\
Portrait4D-v2~\cite{deng2024portrait4dv2}  & 2.084 & 0.071 \\
Ours & 2.138 & 0.089 \\ 
\bottomrule
\end{tabular}
\vspace{-0.5cm}
\end{center}
\caption{\textbf{Additional cross-reenactment metrics.} We report additional metrics for our cross-reenactment evaluation. }
\label{tab:cross_reenactment_quanitative_extra}
\end{table*}

\section{Datasets}
\label{sec:dataset}

We train the MMDM using the monocular video dataset VFHQ~\cite{xie2022vfhq}, and the multi-view datasets Nersemble~\cite{kirschstein2023nersemble}, MEAD~\cite{kaisiyuan2020mead} and Ava-256~\cite{ava256}. For MEAD, we use the sequences with neutral emotions. For Ava-256, we randomly select 20 sequences with 16 camera views for each subject. 
We jointly estimate 3DMM parameters, camera extrinsics and intrinsics using a multi-view face tracker~\cite{taubner2024flowface}.
For Nersemble and Ava-256, we use the available camera calibration. 
We remove frames where less than 95\% of the head is visible.
For VFHQ, we detect and remove videos with scene changes by checking the acceleration of the keypoints detected using the face tracker.
Also, we use MediaPipe~\cite{zhang2020mediapipehandsondevicerealtime} to detect and remove frames containing hands, and we use the face tracker to detect and remove frames containing multiple faces.
We estimate the gaze direction using a gaze estimation model~\cite{abdelrahman2022l2cs} and convert it to the eye rotation of the FLAME model.
In multi-view sequences, we use the most forward-facing view to estimate eye gaze. 
During training, we randomly select $R$ reference images and $G$ target images from all views and frames within a sequence with equal probability. 

\section{Evaluation}


In this section, we provide details on our implementation, additional evaluation metrics, and additional ablations. 

\subsection{Implementation}
We use the same predicted FLAME parameters to evaluate our method and our implementations of FlashAvatar~\cite{xiang2024flashavatar} and GaussianAvatars~\cite{qian2024gaussianavatars}.
The FLAME expression and identity shape parameters are extracted for each set of reference images \cite{taubner2024flowface}. 
Then, we fit expression parameters to the target frames while preserving the identity parameters. 

All metrics for self-reenactment are measured after center-cropping to the head region, resizing to $512 \times 512$ resolution, and removing the background using masks included in the Nersemble dataset.

\subsection{Additional Metrics}
We provide results with the structural similarity metric SSIM, and following previous work~\cite{chu2024gagavatar,kirschstein2024diffusionavatars}, we evaluate the average expression distance (AED) and average pose distance (APD) predicted using DECA~\cite{deca} for both self and cross-reenactment. 
For cross-reenactment, only the jaw pose distance is used.
We measure the average keypoint distance (AKD) using facial landmarks predicted from a keypoint detector~\cite{bulat2017fan}.
We report these additional metrics for self-reenactment with different numbers of reference images in \cref{tab:self_reenactment_combined_extra}, and for cross-reenactment in \cref{tab:cross_reenactment_quanitative_extra}. 
The metrics show that our approach outperforms the baselines for the self-reenactment task. 
For cross-reenactment, our method and the baselines all have similar quantitative scores, but our approach shows clear improvements, as demonstrated in the user study and qualitative results. 

\subsection{User Study}

For each video pair, we ask each participant questions to assess their preference in the following criteria using the following prompts.
\begin{itemize}
    \item \textbf{Visual Quality (VQ):} Evaluate the clarity and visual appeal of each video. Your assessment should focus on the face and head region and ignore the neck and upper body.
    \item \textbf{Expression Transfer (EQ):} Determine which generated avatar's facial expressions better match the driving video.
    \item \textbf{3D Structure (3DS):} Assess how well the 3D structure of the head is preserved across different viewing angles and expressions.
    \item \textbf{Temporal Consistency (TC):} Examine how smoothly and naturally the avatar maintains consistent appearance, expression, and movement across consecutive video frames.
    \item \textbf{Overall Preference (OP):} State your overall preference between the two videos. This is your subjective appraisal of which avatar, in your view, performs better. 
\end{itemize}

The video pairs are presented in a random order, and each participant is asked to select either the left or right video for each criterion. We collect a total number of 4800 responses from 24 participants and conduct $\chi^2$-tests to evaluate statistical significance at the $p < 0.05$ level.







\subsection{Additional Results}
\begin{figure*}
    \begin{center}
    \includegraphics[width=0.85\textwidth]{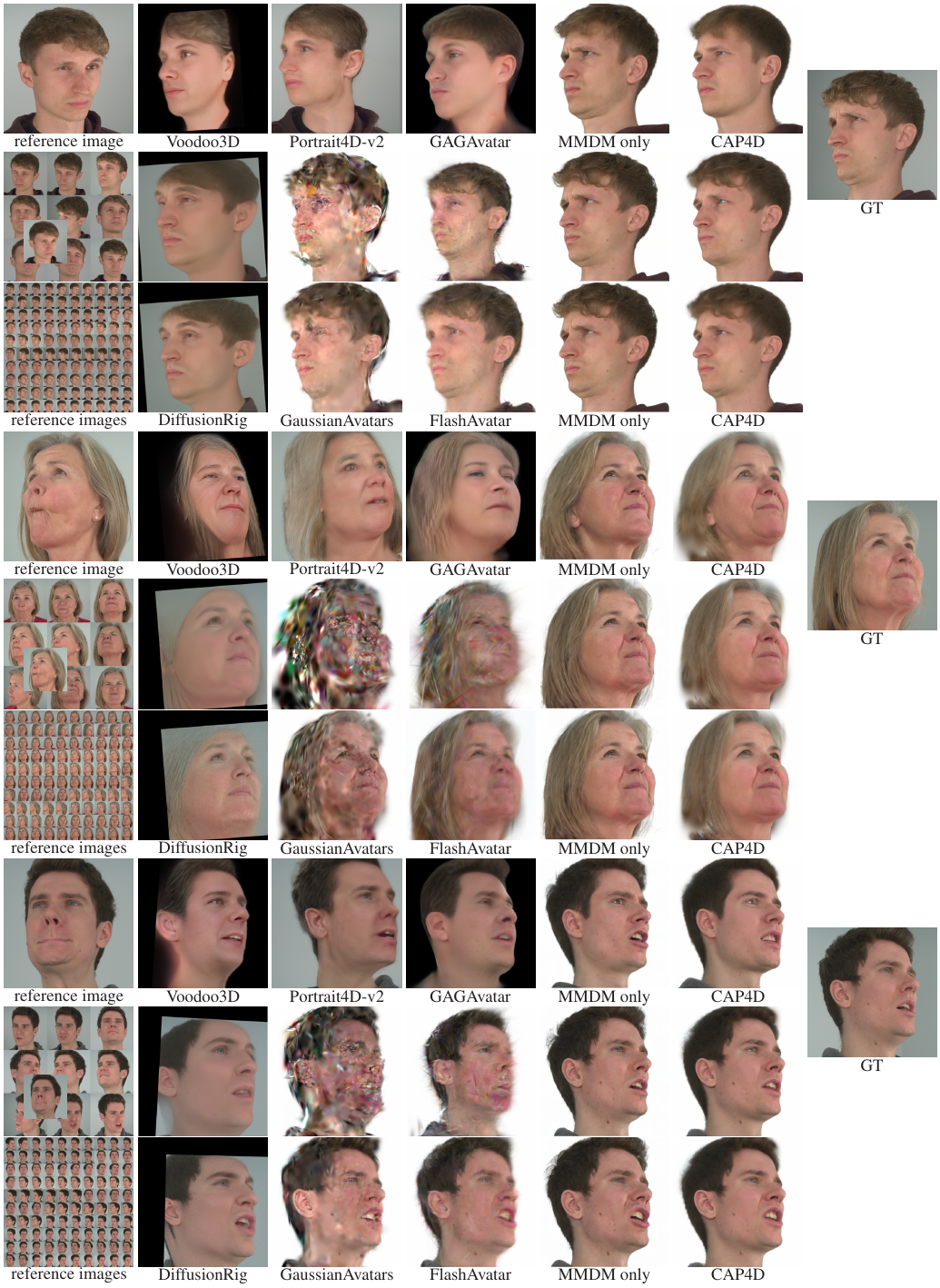}
    \caption{\textbf{Self-reenactment results.} We show more qualitative results for our self-reenactment evaluation with varying numbers of reference frames. Both our MMDM and final 4D avatar can leverage additional reference images to produce details that are not visible in the first reference image (hair, top three rows: birthmarks, last three rows). Our results are significantly better compared to previous methods, especially when the view direction differs greatly from the reference image. }
    \label{fig:self_qualitative_2}
    \end{center}
\end{figure*}

\begin{figure*}
    \includegraphics[width=\textwidth]{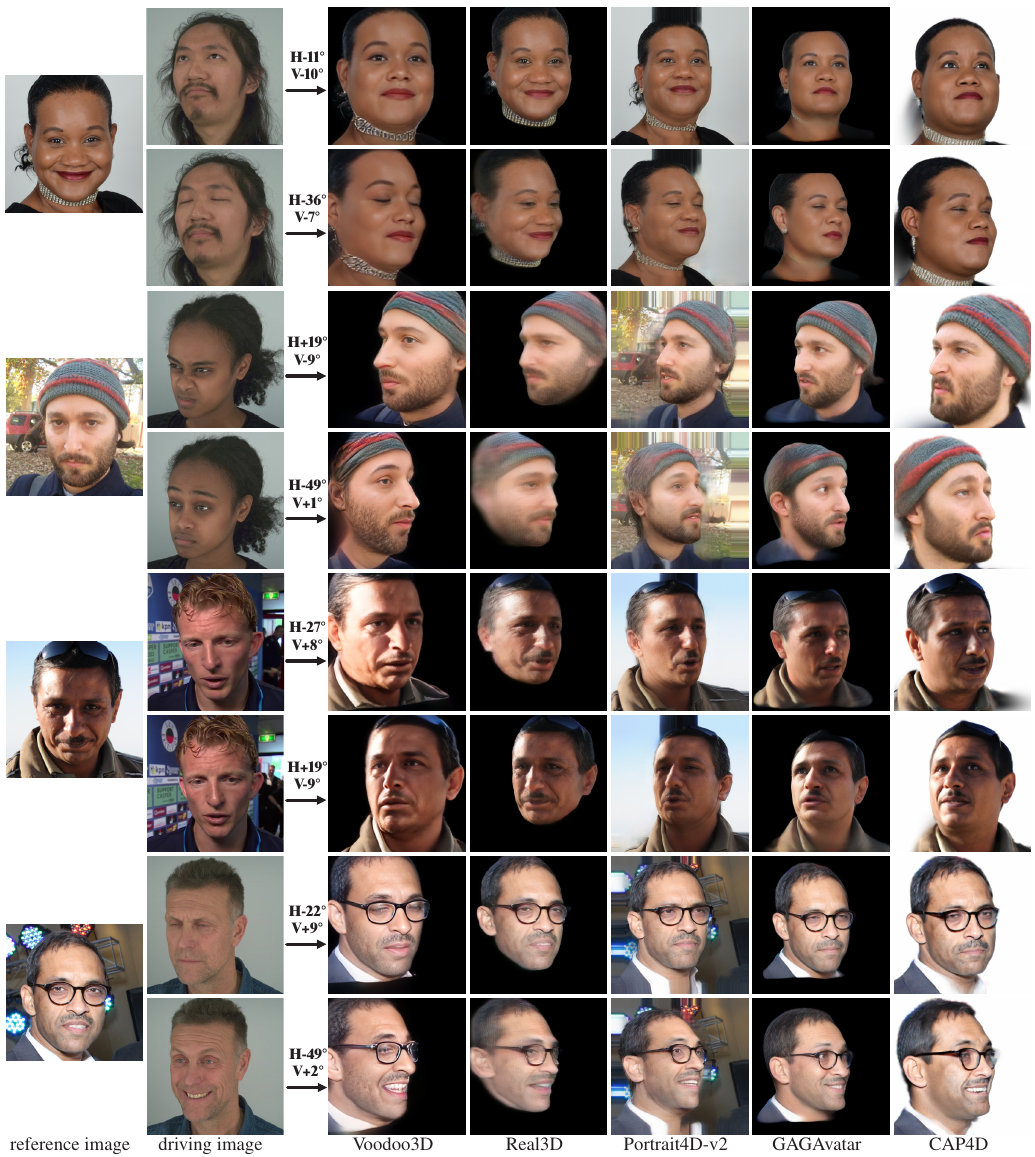}
    \caption{\textbf{Cross-reenactment results.} We show additional qualitative results of our cross-reenactment evaluation. We show generated frames under different driving expressions and viewing angles. Our method consistently produces 4D avatars of higher visual quality and 3D consistency even across challenging view deviations. Our avatar can also model realistic view-dependent lighting changes (row 5 and 6). Best viewed zoomed in. }
    \label{fig:cross_qualitative_2}
\end{figure*}

We provide additional self-reenactment and cross-reenactment results in~\cref{fig:self_qualitative_2,fig:cross_qualitative_2}.
For self-reenactment, we directly compare our method to baselines with one, ten, or 100 reference images. 
As the number of reference images increases, our approach (both MMDM-only and CAP4D) improve in quality. With 100 reference images, we recover fine details in the hair and blemishes on the face.
Our approach better preserves the identity and exhibits higher visual fidelity compared to baselines in the case of one reference image. We improve photorealism compared to baselines using ten or 100 reference images.
The cross-reenactment results show trends similar to those in the main text. 
Compared to baselines, CAP4D better preserves the identity and fine details of the hair, face, and attire.

\clearpage

\begin{figure*}
    \includegraphics[width=\textwidth]{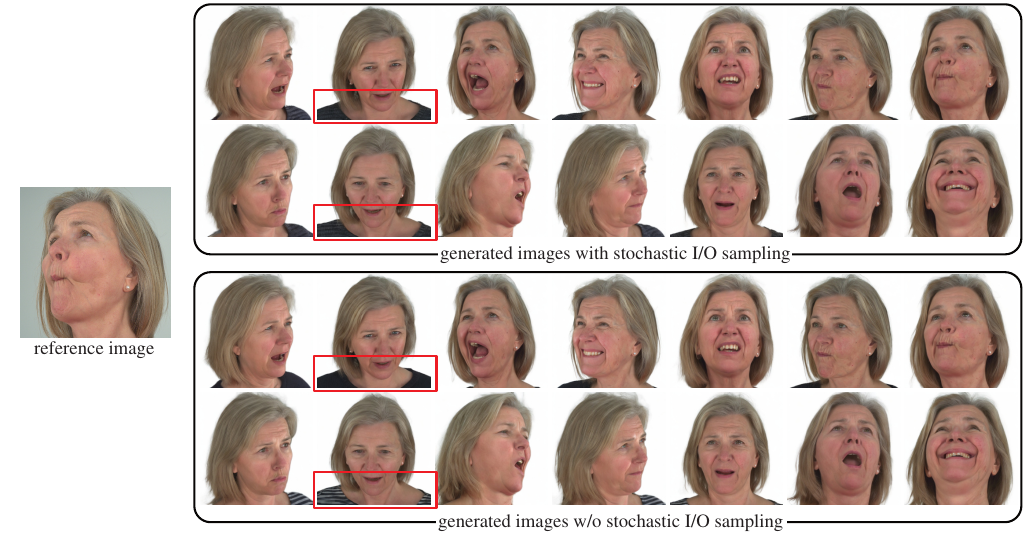}
    \caption{\textbf{Ablations on stochastic I/O sampling.} We generate 14 images from a single reference image (left). Parts of the body do not appear in the reference image (shirt) and are thus ambiguous. When we generate these images without stochastic I/O sampling, two batches of seven images are generated separately (i.e., the third and last rows). This results in visible inconsistencies, such as a changing shirt pattern (red boxes). Stochastic sampling (first and second row) generates these images with information shared across all frames, resulting in a consistent shirt appearance. }
    \label{fig:ablation_sampling}
\end{figure*}
\begin{figure*}
    \includegraphics[width=\textwidth]{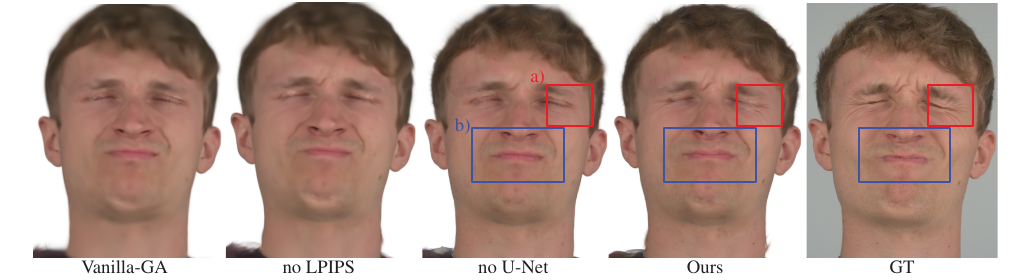}
    \caption{\textbf{Ablations on 4D avatar.} We show qualitative results with ablations of our 4D avatar for self-reenactment, using 10 reference images and images generated using the MMDM. From the left: The original implementation of GaussianAvatars without modifications (Vanilla-GA), our model without LPIPS loss (no LPIPS), our model without the U-Net correcting deformations (no U-Net), our final version (Ours), and the ground truth (GT). Without LPIPS, the avatar appears significantly more blurry. With the U-Net deformations, dynamic details such as wrinkles are depicted more accurately (wrinkles in (a)), and expressions are depicted overall more accurately (lips and nasio-labial fold in (b)). }
    \label{fig:ablation_4d}
\end{figure*}

\clearpage
\clearpage

\begin{table}[h]
\tablefont
\setlength{\tabcolsep}{3pt}
\begin{center}
\begin{tabular}{c l|c c c c} 
\toprule
Category & Ablation & PSNR $\uparrow$ & LPIPS $\downarrow$ & CSIM $\uparrow$ & JOD $\uparrow$ \\

\midrule

\textbf{sampling}       & w/o stochastic & 21.60 & 0.325 & 0.625 & 5.31 \\ 
\textbf{(single ref.)}  & Ours           & 21.82 & 0.317 & 0.632 & 5.40 \\ 

\midrule
\multirow{4}{*}{\textbf{4D rep.}} & w/o stochastic & 21.50 & 0.320 & 0.624 & 5.60 \\ 
&                                   $G=420$        & 21.37 & 0.328 & 0.620 & 5.48 \\ 
&                                   Ours ($G=840$) & 21.69 & 0.311 & 0.633 & 5.67  \\ 
&                                   $G=1260$       & 21.71 & 0.313 & 0.632 & 5.69  \\ 
\bottomrule
\end{tabular}
\end{center}
\vspace{-1em}
\caption{\textbf{Additional ablations.} We assess the impact of our stochastic I/O conditioning with a single reference image. Also, we show the impact of stochastic I/O conditioning, and the number of generated images on the 4D reconstruction. }
\label{tab:ablation_extra}
\end{table}

\subsection{Additional Ablation Study}

\paragraph{Stochastic I/O sampling.}
We conduct experiments with and without the stochastic I/O sampling on the self-reenactment task with a single reference image, and report the results in \cref{tab:ablation_extra}. 
With only a single reference image, stochastic sampling improves image quality (PSNR and LPIPS) and consistency between frames (JOD).

In \cref{fig:ablation_sampling}, we illustrate the improved consistency by generating a toy set of 14 images. 
Inconsistencies such as changing shirt patterns appear without stochastic sampling, whereas using stochastic sampling improves overall consistency.

\paragraph{4D avatar.}
We conduct additional ablations on the ten-reference-image self-reenactment task (see \cref{fig:ablation_4d} and \cref{tab:ablation_extra}); we show the impact of our improvements to the GaussianAvatars representation~\cite{qian2024gaussianavatars} and the impact of stochastic I/O sampling and varying the number of generated images on the quality of the 4D avatar. 
Qualitatively, adding the LPIPS loss and expression-dependent deformations predicted by the U-Net improves the ability to reconstruct wrinkles and expression-dependent details (\cref{fig:ablation_4d}).

The number of generated images also affects the quality of the avatar. Specifically, we evaluate generating $G=420$, 840, or 1260 images and reconstructing the avatar.
When generating fewer images, we observe worse reconstruction (LPIPS); adding additional images beyond $G=840$ does not significantly improve the results, but requires additional compute. 
We also find that stochastic sampling improves the 4D avatar in terms of PSNR, LPIPS, CSIM, and JOD. 

\begin{figure}
    \begin{center}
        \includegraphics[width=0.70\linewidth]{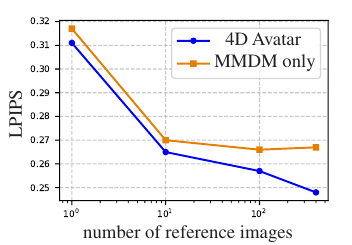}
    \end{center}
    \caption{\textbf{Analysis of Reference Quanitity.} We observe that with hundreds of reference images, the performance of the MMDM plateaus, while our 4D avatar continues scaling with the input quantity. This shows that the 4D avatar can seamlessly benefit from both generated and reference images.  }
    \label{fig:more_reference_images}
\end{figure}

\paragraph{More reference images.}
Finally, we conduct an additional experiment with 400 reference images, which we evaluate the same way as previous experiments on the self-reenactment task. Evaluations (plotted in \cref{fig:more_reference_images}) show that while the performance of our MMDM plateaus, the final 4D avatar can leverage and scale with hundreds of reference images. This is likely because the 4D avatar can improve its reconstruction from the reference images through the direct optimization procedure.






\subsection{Failure Cases}
\begin{figure}
    \includegraphics[width=\linewidth]{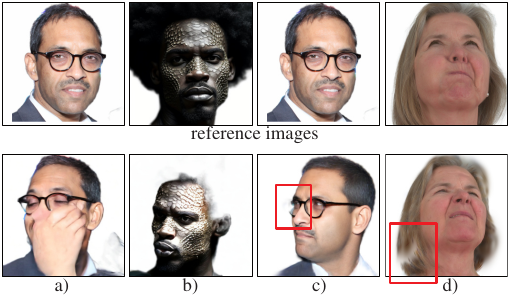}
    \caption{\textbf{Failure Cases.} (a) Our training dataset contains some images that were not properly filtered out using our automated pipeline, and so the MMDM sometimes generates images where the face is occluded (e.g., by a hand). (b) Some training images contain a faulty background segmentation, occasionally leading to artifacts in the  MMDM output. (c) Our 4D avatar can fail in areas not modeled by the FLAME topology, such as glasses and (d) long hair. }
    \label{fig:failure_cases}
\end{figure}

We show failure cases for the MMDM and the 4D avatar in \cref{fig:failure_cases}. 
Specifically, we find that some generated images reflect imperfections in our training dataset, such as images where a hand occludes the face, or images where there are artifacts due to imperfect background segmentation. Also, the 4D avatar cannot model certain regions perfectly, such as hair or glasses, since these are not modeled in the FLAME topology.

\end{document}